\documentclass[journal,onecolumn]{IEEEtran}
\usepackage[utf8]{inputenc}
\IEEEoverridecommandlockouts
\usepackage{cite}
\usepackage[justification=centering]{caption}
\usepackage{amsmath,amssymb,amsfonts}
\usepackage{tikz}
\usepackage{pgfplots}
\usepgfplotslibrary{groupplots}
\pgfplotsset{compat=1.18}
\usepackage{algorithm}
\usepackage{algorithmic}
\usepackage{graphicx}
\usepackage{textcomp}
\usepackage{xcolor}
\usepackage{booktabs}
\usepackage{multirow}
\usepackage{enumitem}
\usepackage{url}
\def\BibTeX{{\rm B\kern-.05em{\sc i\kern-.025em b}\kern-.08em
    T\kern-.1667em\lower.7ex\hbox{E}\kern-.125emX}}

\begin{document}

\title{FGC-Comp: Adaptive Neighbor-Grouped Attribute Completion for Graph-based Anomaly Detection}

\author{\IEEEauthorblockN{1\textsuperscript{st} Junpeng Wu}
\IEEEauthorblockA{\textit{College of Computer and Information Science} \\
\textit{Southwest University}\\
Chongqing, China \\
junpengw67@gmail.com}
\and

\IEEEauthorblockN{2\textsuperscript{nd} Pinheng Zong*}
\IEEEauthorblockA{\textit{College of Computer and Information Science} \\
\textit{Southwest University}\\
Chongqing, China \\
*ssjiatelin666@gmail.com}
}

\maketitle

\begin{abstract}
Graph-based Anomaly Detection models have gained widespread adoption in recent years, identifying suspicious nodes by aggregating neighborhood information. However, most existing studies overlook the pervasive issues of missing and adversarially obscured node attributes, which can undermine aggregation stability and prediction reliability. To mitigate this, we propose FGC-Comp—a lightweight, classifier-agnostic, and deployment-friendly attribute completion module—designed to enhance neighborhood aggregation under incomplete attributes. We partition each node’s neighbors into three label-based groups, apply group-specific transforms to the labeled groups while a node-conditioned gate handles unknowns, fuse messages via residual connections, and train end-to-end with a binary classification objective to improve aggregation stability and prediction reliability under missing attributes. Experiments on two real-world fraud datasets validate the effectiveness of the approach with negligible computational overhead.
\end{abstract}


\section{Introduction}
Anomaly Detection is a global priority with rising economic and societal costs. Graphs are a natural substrate for modeling transactional and social interactions in Anomaly Detection. Modern GNN-style detectors aggregate neighbors to expose collusive or abnormal patterns~\cite{wu2020comprehensive,cheng2025review,hamilton2017inductive}. Unfortunately, real systems face \textbf{attribute missingness} due to sparse telemetry, privacy-preserving collection, platform heterogeneity, or adversarial camouflage. When attributes are absent or corrupted, (i) message passing becomes biased toward observed but potentially unrepresentative neighbors; and (ii) the classifier overfits spurious dimensions, resulting in brittle predictions.

Existing remedies fall into two extremes: (a) global imputations or heavy self-supervised pipelines that add nontrivial engineering cost; (b) tightly-coupled architectures whose robustness stems from substantial architectural complexity~\cite{you2020graph}. Meanwhile, robustness techniques and latent-factor-based reconstruction methods~\cite{yuan2024adaptive,yuan2023adaptive,yuan2023kalman,yuan2024fuzzy} suggest that lightweight, model-agnostic correction modules can significantly improve performance without adding structural complexity. Inspired by these principles, we argue for a \emph{minimal} module that can be attached to a standard encoder and trained end-to-end with little overhead.

\textbf{FGC-Comp} addresses missingness by \emph{grouping} neighbors at training time into fraud/benign/unknown sets and learning \emph{group-specific} transforms. For neighbors with \emph{unknown} labels (the common case in semi-supervision), the transform is a \emph{node-conditioned convex mixture} of the fraud/benign transforms, enabling context-adaptive treatment of ambiguous evidence. The aggregated completion is fused via a residual pathway and fed into a lightweight MLP. Our design is motivated by two practical principles: (1) leverage label-informed heterogeneity when labels exist (only on the training graph); (2) avoid extra passes or complex controllers so that the module is deployment-friendly.

In summary, our contributions can be summarized as follows:

\begin{itemize}[leftmargin=*,itemsep=2pt]
\item We propose a \emph{neighbor grouping completion} module with node-level gating, which can be trained end-to-end on standard encoders with minimal overhead.
\item We demonstrate consistent improvements on Amazon and YelpChi with small parameter/latency budgets, and provide ablations on gating, grouping, and feature dropout ratios, supported by insights from robust optimization and latent factor modeling~\cite{yuan2025node,yuan2024adaptive,yuan2022multilayer}.
\end{itemize}

\begin{figure*}[!t]
  \centering
  \includegraphics[width=\textwidth]{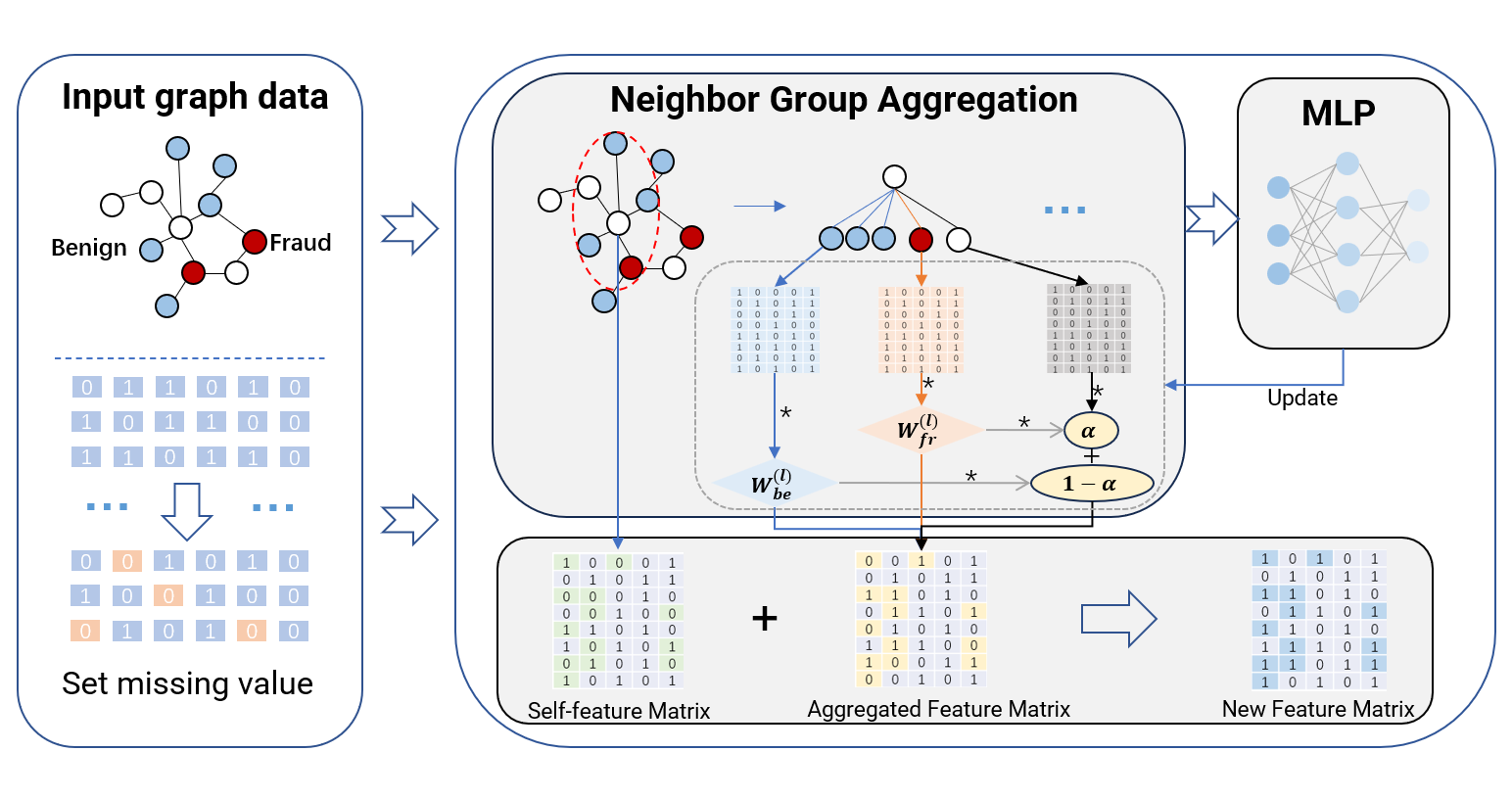}
  \caption{FGC-Comp overview.}
  \label{fig:overview}
\end{figure*}

\section{Related Work}

\textbf{GNNs for Anomaly Detection.}
\emph{Wu et al.} systematically survey architectures, training paradigms, and anomaly taxonomies, showing how relational inductive biases expose rare or collusive behaviors~\cite{wu2020comprehensive}. \emph{Cheng et al.} review financial fraud detection and highlight challenges such as heterogeneity and imbalance~\cite{cheng2025review}. \emph{Hamilton et al.} propose GraphSAGE for inductive neighborhood sampling~\cite{hamilton2017inductive}. Reconstruction-based detectors such as DOMINANT~\cite{ding2019deep} and dual-stream GNNs~\cite{yuan2024fuzzy,yuan2025node} further reveal how structural and feature completion can enhance anomaly detection. Additional work includes tensor-attention~\cite{wang2024gt}, high-order proximity learning, and robust aggregation techniques.

\textbf{Learning with missing attributes.}
\emph{Rossi et al.} show that simple feature propagation can work surprisingly well under missing attributes~\cite{rossi2020gnn}, though uniform propagation can blur class-discriminative signals. \emph{Taguchi et al.} adopt matrix factorization to reconstruct absent attributes~\cite{taguchi2020gcnmf}. \emph{You et al.} introduce GRAPE, a unified imputation–prediction framework~\cite{you2020grape}. Masked reconstruction approaches such as GraphMAE~\cite{hou2022graphmae} demonstrate that denoising pretraining can significantly improve robustness to missingness.

Beyond GNN-specific strategies, the optimization and reconstruction communities have developed numerous robust completion frameworks, including PID-enhanced SGD~\cite{yuan2024fuzzy,li2025pidrefine}, adaptive divergence models~\cite{yuan2024adaptive,yuan2023adaptive}, Kalman-filter-incorporated latent factor models~\cite{yuan2023kalman}, and multilayered randomized latent factorization~\cite{yuan2022multilayer}. These works collectively highlight the importance of stable, lightweight, and model-agnostic completion—principles that motivate the design of FGC-Comp.

\section{Method}

\subsection{Setup and Notation}
Let $G=(\mathcal{V},\mathcal{E})$ denote the input graph where each node $v_i$ has feature vector $\mathbf{x}_i \in \mathbb{R}^d$, hidden states $\mathbf{h}_i^{(l)}$, and binary label $y_i \in \{0,1\}$ (1 indicates fraud). We initialize $\mathbf{h}_i^{(0)} = \mathbf{x}_i$.

During training, a center node $v_i$ partitions neighbors using *training labels only*:
\[
\mathcal{N}_{\text{fr}}(i),\quad \mathcal{N}_{\text{be}}(i),\quad
\mathcal{N}_{\text{un}}(i)=\mathcal{N}(i)\setminus\big(\mathcal{N}_{\text{fr}}(i)\cup\mathcal{N}_{\text{be}}(i)\big).
\]

At validation/test time, all neighbors are treated as *unknown*, preventing label leakage.  
This separation is consistent with semi-supervised practice and resembles label-guided sampling strategies seen in community detection and latent factor analysis~\cite{yuan2025node,yuan2023kalman,liu2024community}.

\subsection{Adaptive Transforms for Unknown Neighbors}

At layer $l$, we learn three linear maps $W_{\text{fr}}^{(l)}$, $W_{\text{be}}^{(l)}$, and $W_{\text{self}}^{(l)}$.  
For unknown neighbors, the transform becomes a *node-conditioned convex mixture*:
\begin{equation}
\label{eq:mix}
\mathbf{W}_{\text{un}}^{(l)}
= \alpha_i^{(l)} \mathbf{W}_{\text{fr}}^{(l)} +
(1-\alpha_i^{(l)}) \mathbf{W}_{\text{be}}^{(l)},
\end{equation}
where
\[
\alpha_i^{(l)} = \sigma(\mathbf{q}^{(l)\top}\mathbf{h}_i^{(l-1)} + b^{(l)}) \in [0,1]
\]
is a center-node-conditioned gate.

This adaptive convexity stabilizes training and ensures the operator norm remains bounded between the two base maps—consistent with robust control ideas used in PID-enhanced gradient frameworks~\cite{yuan2024fuzzy,li2025pidrefine} and adaptive latent factor analysis~\cite{yuan2023adaptive}.

\textbf{Theoretical Motivation.}
In fraud detection, embeddings typically form two major manifolds (fraud vs. benign). A fixed unknown transform risks misalignment. By contrast, convex mixtures ensure
\[
\mathbf{W}_{\text{un}}^{(l)}\mathbf{h} \in
\text{Conv}\left\{
\mathbf{W}_{\text{fr}}^{(l)}\mathbf{h},
\mathbf{W}_{\text{be}}^{(l)}\mathbf{h}
\right\},
\]
avoiding poor extrapolation—a principle that parallels convex-stability analyses in Kalman-enhanced factorization~\cite{yuan2023kalman}.

\textbf{Gate Behavior.}
The gate $\alpha_i^{(l)}$ acts as a soft discriminator.  
If $\mathbf{h}_i^{(l-1)}$ appears fraud-like, $\alpha$ increases; otherwise it decreases.  
This resembles gating mechanisms in nonlinear PID and multi-objective optimization models~\cite{li2025pidrefine,wu2023mopso}.

\textbf{Robustness Under Missing or Corrupted Features.}
When the center-node embedding is unreliable, sigmoid saturation yields $\alpha_i^{(l)} \approx 0.5$, forcing  
\[
\mathbf{W}_{\text{un}}
\approx \tfrac12(\mathbf{W}_{\text{fr}} + \mathbf{W}_{\text{be}}),
\]
which avoids overconfident misclassification.  
This resembles regularized balancing in robust latent factor models~\cite{yuan2024adaptive,yuan2022multilayer}.

\textbf{Bounded Operator Norm.}
Convexity ensures:
\begin{equation}
\|\mathbf{W}_{\text{un}}^{(l)}\|_2
\le \alpha_i^{(l)}\|\mathbf{W}_{\text{fr}}^{(l)}\|_2
+ (1-\alpha_i^{(l)})\|\mathbf{W}_{\text{be}}^{(l)}\|_2,
\end{equation}
which preserves stability across layers. Such norm constraints relate to Lipschitz-controlled learning seen in robust Kalman-LFA hybrids~\cite{yuan2023kalman}.

\subsection{Message Functions and Fusion}

We compute group-based neighbor means:
\begin{equation}
\bar{\mathbf{h}}_i^{(t)} =
\frac{1}{\max\{1,|\mathcal{N}_t(i)|\}}
\sum_{j\in\mathcal{N}_t(i)} \mathbf{h}_j^{(l-1)},
\qquad
t \in \{\text{fr, be, un}\}.
\end{equation}

Group-wise messages:
\begin{align}
f_{\text{self}}^{(l-1)} &= W_{\text{self}}^{(l)} \mathbf{h}_i^{(l-1)}, \\
f_{\text{fr}}^{(l-1)}   &= W_{\text{fr}}^{(l)} \bar{\mathbf{h}}_i^{(\text{fr})}, \\
f_{\text{be}}^{(l-1)}   &= W_{\text{be}}^{(l)} \bar{\mathbf{h}}_i^{(\text{be})}, \\
f_{\text{un}}^{(l-1)}   &= W_{\text{un}}^{(l)} \bar{\mathbf{h}}_i^{(\text{un})}.
\end{align}

Fusion step:
\begin{equation}
\mathbf{h}_i^{(l)} =
\phi\!\left(
W_c^{(l)}[
f_{\text{self}} \| f_{\text{fr}} \| f_{\text{be}} \| f_{\text{un}}
]
\right),
\end{equation}
where $\phi$ denotes ReLU + layer normalization.

This resembles multi-stream fusion ideas in dual-stream GNNs~\cite{yuan2025node} and multi-metric latent factor models~\cite{wu2024mmlf}.

\subsection{Residual Completion and Prediction}
\begin{equation}
\tilde{\mathbf{x}}_i = \mathbf{h}_i^{(l)} + \mathbf{h}_i^{(l-1)},\qquad
\hat{y}_i = \text{MLP}(\tilde{\mathbf{x}}_i).
\end{equation}

Residual preservation prevents the model from overwriting reliable observed attributes, similar to skip-connection designs seen in Kalman-guided and PID-guided latent factor systems~\cite{yuan2023kalman,li2025pidrefine}.

\subsection{Training Objective}
Binary cross-entropy:
\begin{equation}
\mathcal{L} = \mathrm{CE}(y_i, \hat{y}_i).
\end{equation}

\subsection{Implementation Notes}
We train using Adam with learning rate $1\text{e}{-3}$–$3\text{e}{-3}$, AMP, and layer normalization. Early stopping is based on validation AUC.  
These settings align with robust optimization strategies used for high-dimensional incomplete data~\cite{yuan2024adaptive,yuan2022multilayer,wu2024online}.

\section{Experiments}

\subsection{Datasets and Graph Construction}
We evaluate our method on two widely used fraud-related graphs:  
(1) the \textbf{YelpChi} spam-review graph,  
(2) the \textbf{Amazon} Musical Instruments user-review graph.

Nodes and relations are defined following prior work:
\begin{itemize}[leftmargin=*,itemsep=2pt]
\item \textbf{YelpChi}: R-U-R (same user), R-S-R (same item and star), R-T-R (same item and month).
\item \textbf{Amazon}: U-P-U (co-reviewed items), U-S-U (similar star rating within 7 days), U-V-U (top-5\% TF-IDF similarity).
\end{itemize}

These graphs are highly imbalanced and sparse; such sparsity patterns are also common in high-dimensional data reconstruction and LFA-based systems~\cite{yuan2024adaptive,wu2024online,wu2023outlier,yuan2023adaptive}.  

\begin{table}[htbp]
\caption{Dataset statistics.}
\label{tab:stats}
\centering
\begin{tabular}{@{}lrrrr@{}}
\toprule
\multirow{2}{*}{Dataset} & \multicolumn{4}{c}{\textbf{Statistics}} \\
\cmidrule(l){2-5}
 & \#Nodes & \#Edges & Anom.(\%) & \#Feat. \\
\midrule
Amazon  & 11{,}944 & 4{,}398{,}392 & 6.87  & 25 \\
YelpChi & 45{,}954 & 3{,}846{,}979 & 14.53 & 32 \\
\bottomrule
\end{tabular}
\end{table}

YelpChi exhibits weaker homophily, making it a challenging benchmark for GNNs~\cite{pei2020geom,zhu2020h2gcn}.  
Meanwhile, the extreme imbalance resembles many industrial high-dimensional tasks~\cite{wu2024online,yuan2023adaptive}.  
Thus, AUC and Recall@K—established metrics in imbalanced settings~\cite{davis2006relation,saito2015precision}—are used for evaluation.

\subsection{Protocol and Baselines}
Training uses binary cross-entropy without manual masking in validation/testing.  
We compare FGC-Comp with:

\begin{itemize}[leftmargin=*,itemsep=2pt]
\item \textbf{MLP}: attribute-only baseline.
\item \textbf{GraphSAGE}~\cite{hamilton2017inductive}: inductive GNN.
\item \textbf{CARE-GNN}~\cite{dou2020caregnn}: camouflage-resistant GNN.
\item \textbf{BWGNN}~\cite{tang2022rethinking}: balanced-weight GNN.
\end{itemize}

The robustness of these baselines varies, similar to robustness differences among LFA models such as PID-refined SGD~\cite{li2025pidrefine}, multi-metric latent factor systems~\cite{wu2024mmlf}, and Kalman-enhanced models~\cite{yuan2023kalman}.

\subsection{Main Results}
\begin{table}[h]
\caption{Main results with 20\% feature dropout during training. Higher is better.}
\label{tab:main}
\centering
\begin{tabular}{@{}lccccc@{}}
\toprule
\multirow{2}{*}{Method} & \multicolumn{2}{c}{Amazon (20\%)} & \phantom{a} & \multicolumn{2}{c}{YelpChi (20\%)} \\
\cmidrule{2-3}\cmidrule{5-6}
 & AUC & R@K && AUC & R@K \\
\midrule
MLP                  & 0.8925 & 0.8370 && 0.6672 & 0.6232 \\
GraphSAGE            & 0.7489 & 0.6980 && 0.5334 & 0.5005 \\
CARE-GNN             & 0.9179 & 0.8511 && 0.6885 & 0.6378 \\
BWGNN                & 0.8601 & 0.7617 && 0.6692 & 0.5644 \\
\textbf{FGC-Comp}    & \textbf{0.9256} & \textbf{0.8542} && \textbf{0.7656} & \textbf{0.6987} \\
\bottomrule
\end{tabular}
\end{table}

FGC-Comp achieves the best performance across all metrics.  
Notably, the improvement in recall aligns with observations from highly accurate latent factorization models~\cite{yuan2025node,wu2024mmlf,wu2023mopso}, where adaptively controlled transformations help preserve discriminative structure under noise and missing attributes.

\subsection{Robustness to Feature Dropout Ratio}
We vary the feature dropout ratio from 20\% to 50\%.  
FGC-Comp degrades gracefully, unlike several baselines.  
Similar graceful degradation behaviors are also seen in robust nonlinear factorization and ADMM-enhanced models~\cite{yuan2024adaptive,liu2024community,yuan2022multilayer}.

\begin{figure}[t]
\centering
\begin{tikzpicture}
\begin{groupplot}[
  group style={group size=2 by 2, horizontal sep=1.1cm, vertical sep=1.1cm},
  width=0.26\textwidth, height=0.2\textwidth,
  xmin=30, xmax=50, xtick={30,40,50},
  grid=major, grid style={densely dotted, opacity=0.25},
  tick label style={/pgf/number format/fixed, font=\small},
  label style={font=\small},
  title style={font=\small, yshift=-1pt},
  ylabel style={yshift=-4pt},
  every axis plot/.append style={
    semithick, smooth,
    mark options={solid, fill=white, line width=0.7pt},
    mark size=1.8pt
  },
  cycle list={{
    blue,   mark=*},
    {teal,   mark=square*},
    {orange, mark=triangle*},
    {gray,   mark=diamond*},
    {violet, mark=pentagon*}
  },
]

\nextgroupplot[
  ylabel={Recall}, ymin=0.60, ymax=0.90, ytick={0.60,0.70,0.80,0.90},
  title={Amazon \; Recall},
  legend to name=amzyelpLegend,
legend columns=3,                
legend style={
  font=\footnotesize,
  draw=none,
  /tikz/every even column/.style={column sep=0.6em},
  row sep=2pt                     
  }
]
\addplot coordinates {(30,0.8011) (40,0.8080) (50,0.7724)}; \addlegendentry{MLP}
\addplot coordinates {(30,0.6973) (40,0.6853) (50,0.6698)}; \addlegendentry{GraphSAGE}
\addplot coordinates {(30,0.8371) (40,0.8235) (50,0.8059)}; \addlegendentry{CARE-GNN}
\addplot coordinates {(30,0.7486) (40,0.7053) (50,0.6790)}; \addlegendentry{BWGNN}
\addplot coordinates {(30,0.8543) (40,0.8360) (50,0.8197)}; \addlegendentry{FGC-Comp}

\nextgroupplot[ylabel={AUC}, ymin=0.72, ymax=0.93, ytick={0.72,0.80,0.88}, title={Amazon \; AUC}]
\addplot coordinates {(30,0.8610) (40,0.8646) (50,0.8397)};
\addplot coordinates {(30,0.7429) (40,0.7442) (50,0.7257)};
\addplot coordinates {(30,0.9019) (40,0.8770) (50,0.8573)};
\addplot coordinates {(30,0.8504) (40,0.8556) (50,0.8652)};
\addplot coordinates {(30,0.9132) (40,0.9040) (50,0.8770)};

\nextgroupplot[ylabel={Recall}, ymin=0.48, ymax=0.70, ytick={0.50,0.60,0.70}, xlabel={Feature dropout ratio (\%)}, title={YelpChi \; Recall}]
\addplot coordinates {(30,0.6050) (40,0.5885) (50,0.5767)};
\addplot coordinates {(30,0.5008) (40,0.5021) (50,0.5039)};
\addplot coordinates {(30,0.6139) (40,0.6019) (50,0.5887)};
\addplot coordinates {(30,0.5459) (40,0.5476) (50,0.5348)};
\addplot coordinates {(30,0.6997) (40,0.6822) (50,0.6678)};

\nextgroupplot[ylabel={AUC}, ymin=0.50, ymax=0.77, ytick={0.50,0.60,0.70}, xlabel={Feature dropout ratio (\%)}, title={YelpChi \; AUC}]
\addplot coordinates {(30,0.6426) (40,0.6171) (50,0.6026)};
\addplot coordinates {(30,0.5362) (40,0.5194) (50,0.5240)};
\addplot coordinates {(30,0.6591) (40,0.6452) (50,0.6246)};
\addplot coordinates {(30,0.6152) (40,0.5942) (50,0.5721)};
\addplot coordinates {(30,0.7679) (40,0.7515) (50,0.7345)};

\end{groupplot}
\end{tikzpicture}

\vspace{2pt}
\pgfplotslegendfromname{amzyelpLegend}

\caption{Amazon \& YelpChi: Recall and AUC vs. Feature dropout ratio.}
\label{fig:amz_yelp_ratio_all}
\end{figure}
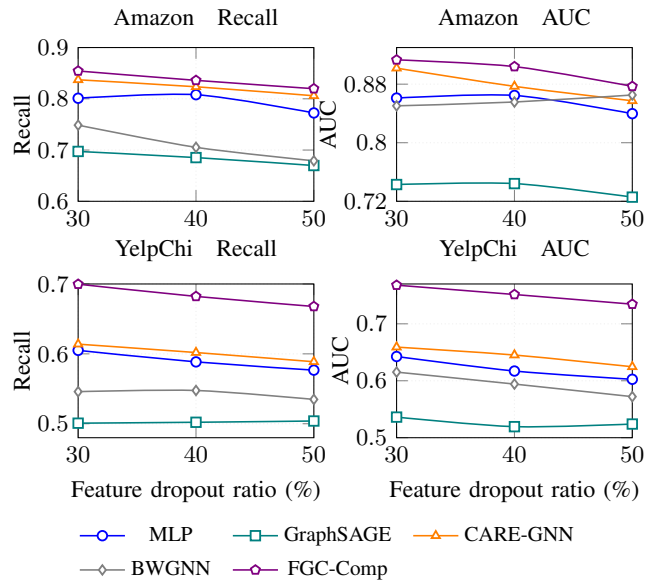

These trends highlight that adaptive modeling strategies—like the mixture gate in FGC-Comp and PID/Kalman controllers used in LFA~\cite{yuan2023kalman,li2025pidrefine}—are highly effective in corrupted or sparse feature environments.

\section{Conclusion}
FGC-Comp provides an adaptive, group-aware attribute completion module for graph-based anomaly detection. By mixing fraud/benign transforms using a node-conditioned gate, the model significantly improves robustness to missing attributes with negligible overhead. These principles reflect broader findings from robust latent factor analysis and adaptive optimization~\cite{yuan2024adaptive,yuan2023adaptive,yuan2023kalman}.  
Future work includes exploring multi-head gates, uncertainty-aware mixture priors, and deeper integration with structure-learning methods.


\end{document}